\documentclass[10pt,twocolumn,letterpaper]{article}

\usepackage{cvpr}
\usepackage{times}
\usepackage{epsfig}
\usepackage{caption, subcaption}
\usepackage{graphicx}
\usepackage{tabularx}
\usepackage{booktabs}
\usepackage{amsmath}
\usepackage{amssymb}
\usepackage{xcolor}
\usepackage{array}
\usepackage{esvect}
\usepackage{mathabx}

\definecolor{darkpastelgreen}{rgb}{0.01, 0.75, 0.24}

\newcommand{\super}[1]{{\color{blue} #1}}
\newcommand{\unsuper}[1]{{\color{red} #1}}

\usepackage[pagebackref=true,breaklinks=true,letterpaper=true,colorlinks,bookmarks=false]{hyperref}

\cvprfinalcopy 

\def\cvprPaperID{838} 

\ifcvprfinal\fi

\newcommand{\darkgrayed}[1]{\textcolor{darkgray}{#1}}
\makeatletter
\newcommand*\titleheader[1]{\gdef\@titleheader{#1}}
\AtBeginDocument{%
  \let\st@red@title\@title
  \def\@title{%
    \vskip-3em
    \bgroup\normalfont\large\centering\@titleheader\par\egroup
    \vskip1.5em\st@red@title}
}
\makeatother

\titleheader{\small \darkgrayed{This paper has been accepted for publication at the IEEE Conference on Computer Vision and Pattern Recognition (CVPR),\\ Long Beach (CA), 2019.
\copyright IEEE}}

\title{Unsupervised Moving Object Detection via \\
Contextual Information Separation}

\begin{document}

\author{Yanchao Yang*\\
UCLA Vision Lab\\
\and
Antonio Loquercio*\\
University of Zurich\\
\and
Davide Scaramuzza\\
University of Zurich\\
\and
Stefano Soatto\\
UCLA Vision Lab\\
}

\maketitle
\thispagestyle{empty}

\begin{abstract}
We propose an adversarial contextual model for detecting moving objects in images. A deep neural network is trained to predict the optical flow in a region using information from everywhere else but that region (context), while another network attempts to make such context as uninformative as possible. The result is a model where hypotheses naturally compete with no need for explicit regularization or hyper-parameter tuning. Although our method requires no supervision whatsoever, it outperforms several methods that are pre-trained on large annotated datasets. Our model can be thought of as a generalization of classical variational generative region-based segmentation, but in a way that avoids explicit regularization or solution of partial differential equations at run-time. We publicly release all our code and trained networks.\footnote{\scriptsize \url{http://rpg.ifi.uzh.ch/unsupervised_detection.html}} \let\thefootnote\relax\footnote{*These two authors contributed equally. Correspondence to {\tt yanchao.yang@cs.ucla.edu} and {\tt loquercio@ifi.uzh.ch} }\addtocounter{footnote}{-1}\let\thefootnote\svthefootnote 

\end{abstract}


\section{Introduction}

\begin{figure}
\centering
\renewcommand{\arraystretch}{0}
\begin{tabular}{@{\hspace{0.0cm}} c @{\hspace{0cm}} c @{\hspace{0cm}} c @{\hspace{0cm}} c@ {\hspace{0cm}} }
    \includegraphics[width=0.25\linewidth]{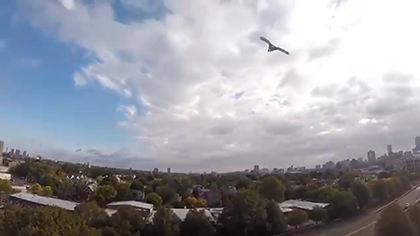} &
    \includegraphics[width=0.25\linewidth]{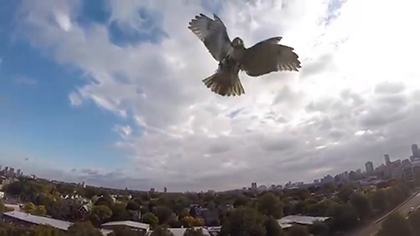} &
    \includegraphics[width=0.25\linewidth]{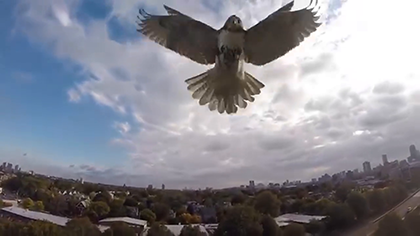} &
    \includegraphics[width=0.25\linewidth]{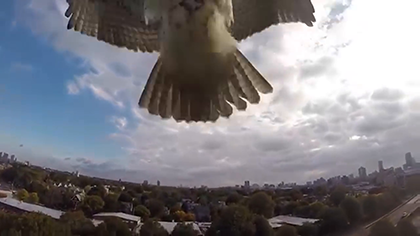} \\
    \includegraphics[width=0.25\linewidth]{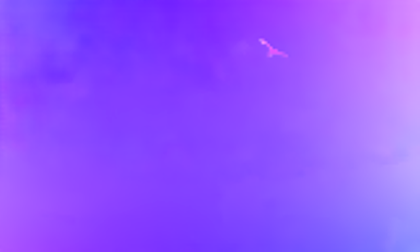} &
    \includegraphics[width=0.25\linewidth]{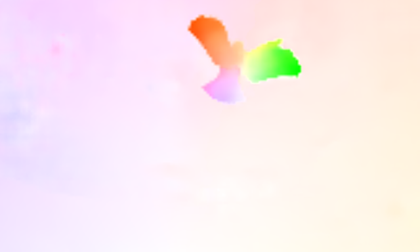} &
    \includegraphics[width=0.25\linewidth]{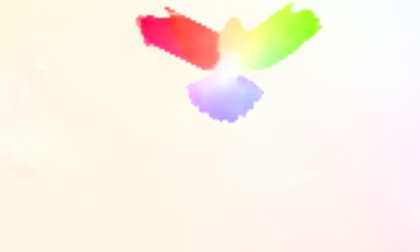} &
    \includegraphics[width=0.25\linewidth]{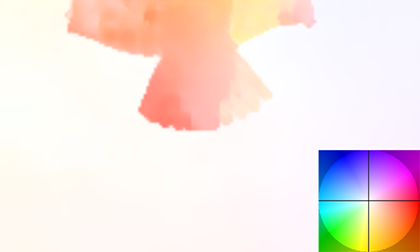} \\ [2mm]
        \includegraphics[width=0.25\linewidth]{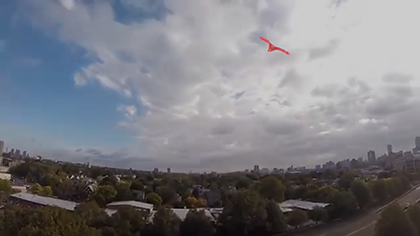} &
    \includegraphics[width=0.25\linewidth]{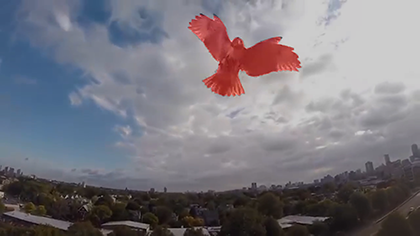} &
    \includegraphics[width=0.25\linewidth]{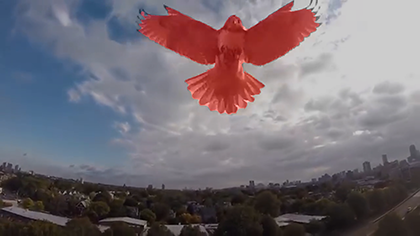} &
    \includegraphics[width=0.25\linewidth]{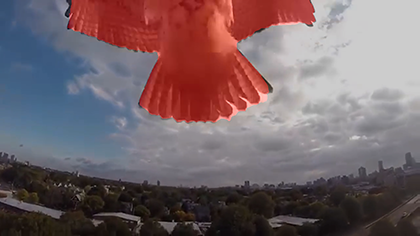} \\ 

\end{tabular}

\caption{\small An encounter between a hawk and a drone (top). The latter will not survive without being aware of the attack. Detecting  moving objects is crucial to the survival of animal and artificial systems alike. Note that the optical flow (middle row) is quite diverse within the region where the hawk projects: It changes both in space and time. Grouping this into a moving object (bottom row) is our goal in this work. Note the object is detected by our algorithm across multiple scales, partial occlusions from the viewpoint, and complex boundaries. }
\vspace{-3mm}
\label{fig:example}
\end{figure}

Consider Fig.~\ref{fig:example}: Even relatively simple objects, when moving in the scene, cause complex discontinuous changes in the image. Being able to rapidly detect independently moving objects in a wide variety of scenes from images is functional to the survival of animals and autonomous vehicles alike. We wish to endow artificial systems with similar capabilities, without the need to pre-condition or learn similar-looking backgrounds. This problem relates to motion segmentation, foreground/background separation, visual attention, video object segmentation as we discuss in Sect.~\ref{sect-relations}. For now, we use the words ``object'' or ``foreground'' informally\footnote{The precise meaning of these terms will be formalized in Sect.~\ref{sect-method}.} to mean (possibly multiple) connected regions of the image domain, to be distinguished from their surrounding, which we call ``background'' or ``context,'' according to {\em some} criterion.

Since objects exist in the scene, not in the image, a method to infer them from the latter rests on an operational definition based on measurable image correlates. We call moving objects regions of the image whose motion cannot be explained by that of their surroundings. In other words, the motion of the background is uninformative of the motion of the foreground and vice-versa. The ``information separation'' can be quantified by the information reduction rate (IRR) between the two as defined in Sect.~\ref{sect-method}. This naturally translates into an adversarial inference criterion that has close connections with classical variational region-based segmentation, but with a twist: Instead of learning a generative model of a region that explains the image {\em in that region} as well as possible, our approach yields a model that tries to explain it {\em as poorly as possible} using measurements from {\em everywhere else but} that region.

In generative model-based segmentation, one can always explain the image with a trivial model, the image itself. To avoid that, one has to impose model complexity bounds, bottlenecks or regularization. Our model does not have access to trivial solutions, as it is forced to predict a region without looking at it. What we learn instead is a contextual adversarial model, without the need for explicit regularization, where foreground and background hypotheses compete to explain the data with no pre-training nor (hyper)parameter selection. In this sense, our approach relates to adversarial learning and self-supervision as discussed in Sect.~\ref{sect-relations}.

The result is a completely unsupervised method, unlike many recent approaches that are called unsupervised but still require supervised pre-training on massive labeled datasets and can perform poorly in contexts that are not well represented in the training set. Despite the complete lack of supervision, our method performs competitively even compared with those that use supervised pre-training (Sect.~\ref{sect-experiments}).

\subsection*{Summary of Contributions}

Our method captures the desirable features of variational region-based segmentation: Robustness, lack of thresholds or tunable parameters, no need for training. However, it does not require solving a partial differential equation (PDE) at run-time, nor to pick regularizers or Lagrange multipliers, nor to restrict the model to one that is simple-enough to be tractable analytically. It also exploits the power of modern deep learning methods: It uses deep neural networks as the model class, optimizes it efficiently with stochastic gradient descent (SGD), and can be computed efficiently at run time. However, it requires no supervision whatsoever.

\def\u{u} 
\def\ui{{\u^{\tiny \rm in}}}
\def\uo{{\u^{\tiny \rm out}}}
\def\ux{{\u_\x}}
\def\uix{{\ux}^{\tiny \rm in}}
\def\uox{{\ux}^{\tiny \rm out}}
\def\I{I} 
\def\x{i}
\def\D{D} 
\def\o{\Omega} 
\def\y{j}
\def\uy{\u_j}
\def\uoy{{\u_j}^{\rm \tiny out}}
\def\uiy{{\u_j}^{\rm \tiny in}}

\begin{figure*}
    \centering
    \includegraphics[width=0.75\textwidth]{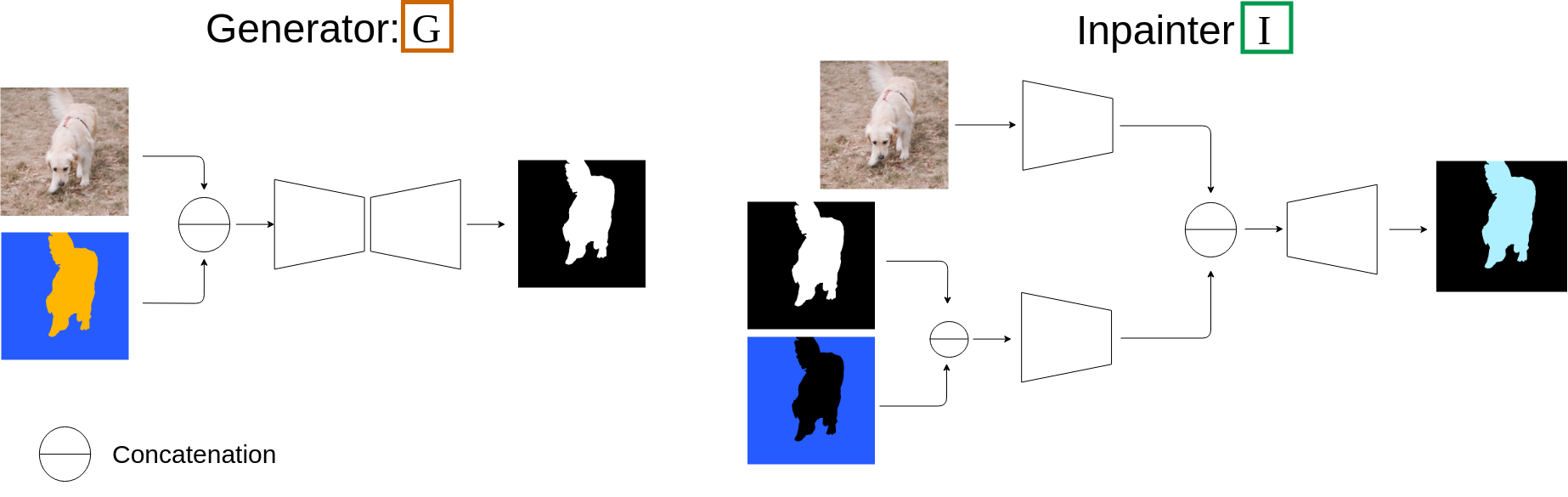}
    \caption{\small During training, our method entails two modules. One is the generator (G) which produces a mask of the object by looking at the image and the associated optical flow. The other module is the inpainter (I) which tries to inpaint back the optical flow masked out by the corresponding mask. Both modules employ the encoder-decoder structure with skip connections. However, the inpainter (I) is equipped with two separate encoding branches. See Sect.~\ref{sect-implementation} for network details.}
    \label{fig:modules}
    \vspace{-2mm}
\end{figure*}

\begin{figure*}
    \centering
    \includegraphics[width=0.85\textwidth]{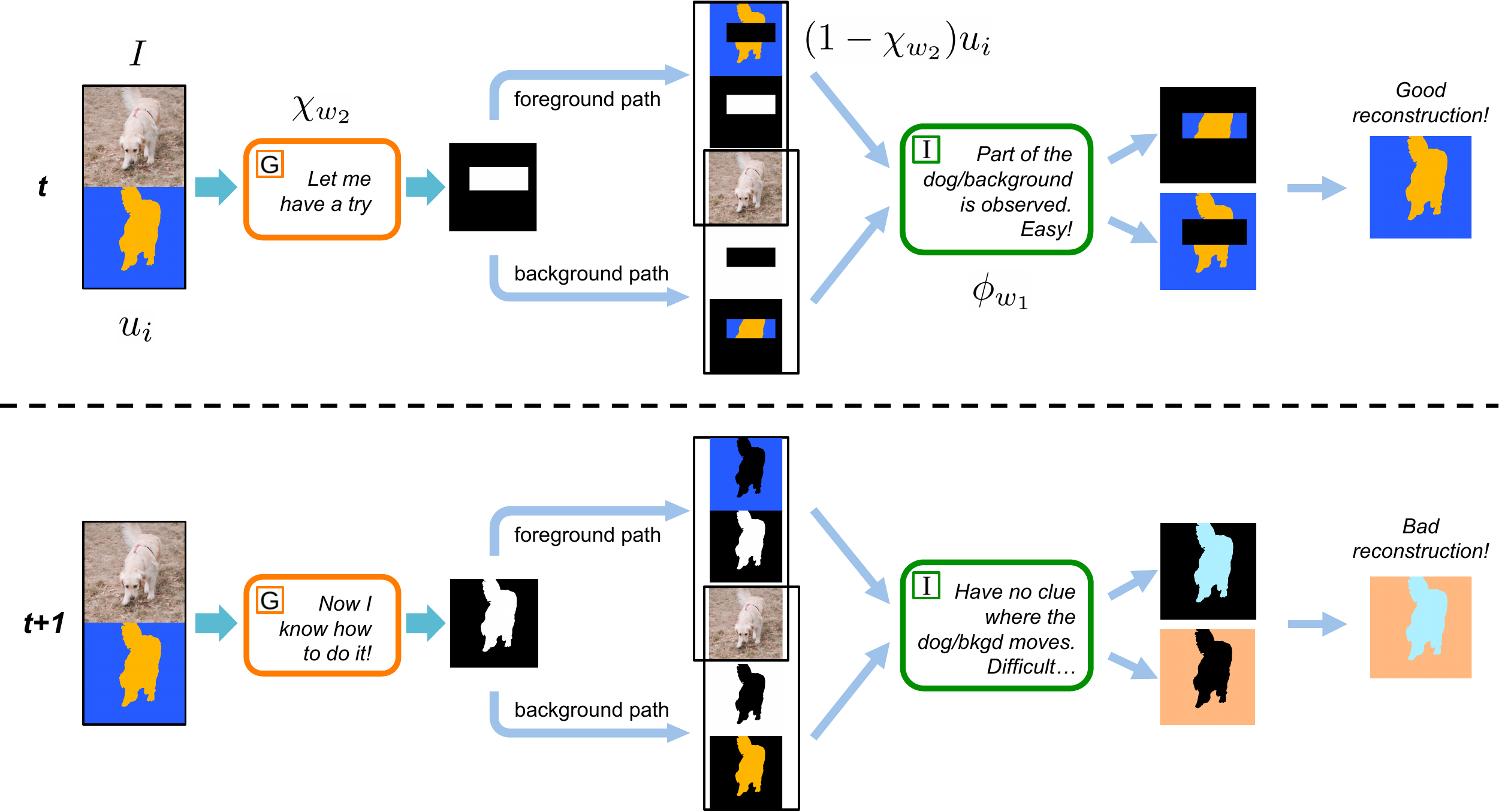}
    \caption{\small The two diagrams illustrate the learning process of the mask generator (G), after the inpainter (I) has learned how to accurately inpaint a masked flow.
    The upper diagram shows a poorly trained mask generator which does not precisely detect the object.
    Due to the imprecise detection, the inpainter can observe part of the object's flow, and perform an accurate reconstruction.
    At the same time, the inpainter partially observes the background's flow in the complementary mask. Consequently, it can precisely predict missing parts of the background's flow.
    In contrast, the lower diagram shows a fully trained mask generator which can precisely tell apart the object from the background.
    In this case, the inpainter observes the flow only outside the object and has no information to predict the flow inside it.
    At initialization time the inpainter does not know the conditionals to inpaint masked flows.
    Therefore, we propose to train both the generator and the inpainter jointly  in an adversarial manner (see Sect.~\ref{sect-method}).}
    \label{fig:method}
      \vspace{-2mm}
\end{figure*}
While our approach has close relations to both classical region-based variational segmentation and generative models, as well as modern deep learning-based self-supervision, discussed in detail in Sect.~\ref{sect-relations}, to the best of our knowledge, it is the first {\em adversarial contextual model} to detect moving objects in images.
It achieves better or similar performance compare to unsupervised methods on the three most common benchmarks, and it even edges out methods that rely on supervised pre-training, as described in Sect.~\ref{sect-experiments}. On one of the considered benchmarks, it outperforms all methods using supervision, which illustrates the generalizability of our approach.
In Sect.~\ref{sect-discussion} we describe typical failure modes and discuss limitations of our method.

\section{Method}
\label{sect-method}

We call ``moving object(s)'' or ``foreground'' any region of an image whose motion is unexplainable from the context. A ``region of an image'' $\o$ is a compact and multiply-connected subset of the domain of the image, discretized into a lattice $\D$. ``Context'' or ``background'' is the complement of the foreground in the image domain, $\o^c = \D \setminus \o$. Given a measured image $\I$ and/or its optical flow to the next (or previous) image $\u$, foreground and background are uncertain, and therefore treated as random variables. A random variable $\u_1$ is ``unexplainable'' from (or ``uninformed'' by) another $\u_2$ if their mutual information ${\mathbb I}(\u_1; \u_2)$ is zero, that is if their joint distribution equals the product of the marginals, $P(\u_1, \u_2) = P(\u_1)P(\u_2)$.

More specifically, the optical flow $\u: \D_1 \rightarrow \mathbb{R}^2$ maps the domain of an image $\I_1: \D_1 \rightarrow {\mathbb R}_+^3$ onto the domain $\D_2$ of $\I_2$, so that if $x_i \in \D_1$, then $x_i + \ux \in \D_2$, where $\ux = \u(x_i)$ up to a discretization into the lattice and cropping of the boundary. Ideally, if the brightness constancy constraint equation that defines optical flow was satisfied, we would have $\I_1 = \I_2 \circ \u$ point-wise.

If we consider the flow at two locations $i, j$, we can formalize the notion of foreground as a region $\o$ that is uninformed by the background: 
\begin{align}
\begin{cases}
\mathbb{I}(\u_i, \u_j | \I) > 0, i, j \in \o\\
\mathbb{I}(\u_i, \u_j | \I) = 0, i \in \o, j \in D \setminus \o.
\end{cases}
\label{eq:define-object}
\end{align}
As one would expect, based on this definition, if the domain of an object is included in another, then they inform each other (see appendix \cite{yang2019unsupervised}).

\subsection{Loss function}

We now operationalize the definition of foreground into a criterion to infer it. We use the information reduction rate (IRR) $\gamma$, which takes two subsets $\boldsymbol{x}, \boldsymbol{y} \subset D$ as input and returns a non-negative scalar:
\begin{equation}
\gamma(\boldsymbol{x} | \boldsymbol{y}; \I) = \dfrac{\mathbb{I}(\u_{\boldsymbol{x}}, \u_{\boldsymbol{y}} | \I)}{\mathbb{H}(\u_{\boldsymbol{x}} | \I)} 
= 1 - \dfrac{\mathbb{H}(\u_{\boldsymbol{x}} | \u_{\boldsymbol{y}}, \I)}{\mathbb{H}(\u_{\boldsymbol{x}} | \I)}
\label{def:reduction-rate}
\end{equation}
where $\mathbb{H}$ denotes (Shannon) entropy. It is zero when the two variables are independent, but the normalization prevents the trivial solution (empty set).\footnote{A small constant $0 < \epsilon \ll 1$ is added to the denominator to avoid singularities, and whenever $\boldsymbol{x} \neq \emptyset$, $\mathbb{H}(\u_{\boldsymbol{x}} | \I) \gg \epsilon$, thus we will omit $\epsilon$ from now on.} As proven in the appendix \cite{yang2019unsupervised}, objects as we defined them are the regions that minimize the following loss function
\begin{equation}
\mathcal{L}(\o; \I) = \gamma(\o| \o^c; \I) + \gamma(\o^c |\o; \I).
\label{eq:loss-object}
\end{equation}
Note that $\cal L$ {\em does not have a complexity term}, or regularizer, as one would expect in most region-based segmentation methods. This is a key strength of our approach, that involves no modeling hyperparameters, as we elaborate on in Sect.~\ref{sect-relations}.

Tame as it may look, \eqref{eq:loss-object} is intractable in general. For simplicity we indicate the flow inside the region(s) $\o$ (foreground) with $\ui = \{\ux, \ \x \in \o\} $, and similarly for $\uo$, the flow in the background $\o^c$.
The only term that matters in the IRR is the ratio $\mathbb{H}(\ui | \uo, \I)/\mathbb{H}(\ui | \I)$, which is 
\begin{equation}
\dfrac{\int \log P( \ui | \uo, \I ) dP( \ui | \uo, \I ) }
{\int \log P(\ui | \I) dP(\ui | \I)}
\label{eq:quot-of-entropy}
\end{equation}
that measures the information transfer from the background to the foreground. This is minimized when knowledge of the background flow is sufficient to predict the foreground. To enable computation, we have to make draconian, yet common, assumptions on the underlying probability model, namely that
\begin{eqnarray}
P(\ui=x | \I) &\propto& \exp\left(-\dfrac{\Vert x \Vert^2}{\sigma^2}\right) \label{eq:probability model} \\
P(\ui=x | \uo=y, \I ) &\propto& \exp\left(-\dfrac{\Vert x - \phi(\o, y, \I) \Vert^2}{\sigma^2}\right) \nonumber
\end{eqnarray}
where $
\phi(\o, y, \I) = \int \ui dP(\ui | \uo, \I)
$
is the conditional mean given the image and the complementary observation. Here we assume $\phi(\o, \emptyset, \I) = 0$, since given a single image the most probable guess of the flow is zeros. With these assumptions, \eqref{eq:quot-of-entropy} can be simplified, to
\begin{multline}
\dfrac{\int \Vert \ui - \phi (\o, \uo, \I) \Vert^2 dP( \ui| \uo, \I) }
{\int \Vert \ui \Vert^2 dP(\ui | \I)} \approx \\
\approx \dfrac{\sum_{i=1}^{N}\Vert {\ux}^{\tiny \rm in} - \phi(\o, {\ux}^{\tiny \rm out}, \I) \Vert^2}{\sum_{i=1}^{N}\Vert {\ux}^{\tiny \rm in} \Vert^2}
\end{multline}
where $N = |{\cal D}|$ is the cardinality of ${\cal D}$, or the number of flow samples available. Finally, our loss \eqref{eq:loss-object} to be minimized can be approximated as
\begin{multline}
\mathcal{L}(\o; \I) = 1-\dfrac{\sum_{i=1}^{N} \Vert \uix - \phi(\o, \uox, \I) \Vert^2}{\sum_{i=1}^{N}\Vert \uix \Vert^2 + \epsilon}\\
+ 1 - \dfrac{\sum_{i=1}^{N}\Vert \uox - \phi(\o^c, \uix, \I) \Vert^2}{\sum_{i=1}^{N}\Vert \uox \Vert^2 + \epsilon}.
\label{eq:empirical-symmetric-loss0}
\end{multline}
In order to minimize this loss, we have to choose a representation for the unknown region $\Omega$ and for the function $\phi$.

\subsection{Function class}

The region $\o$ that minimizes \eqref{eq:empirical-symmetric-loss0} belongs to the power set of $D$, that is the set of all possible subsets of the image domain, which has exponential complexity.\footnote{In the continuum, it belongs to the infinite-dimensional set of compact and multiply-connected regions of the unit square.} We represent it with the indicator function
\begin{eqnarray}
    \chi: D &\rightarrow& \{0, 1\} \nonumber \\ 
    \x &\mapsto& 1 \ {\rm if} \ \x \in \o; \ 0 \ {\rm otherwise}
\end{eqnarray}
so that the flow inside the region $\o$ can be written as $\uix = \chi\ux$, and outside as $\uox = (1-\chi)\ux$.

Similarly, the function $\phi$ is non-linear, non-local, and high-dimensional, as it has to predict the flow in a region of the image of varying size and shape, given the flow in a different region. In other words, $\phi$ has to capture the context of a region to {\em recover} its flow.

Characteristically for the ages, we choose both $\phi$ and $\chi$ to be in the parametric function class of deep convolutional neural networks, as shown in Fig. \ref{fig:modules}, the specifics of which are in Sect.~\ref{sect-implementation}. We indicate the parameters with $w$, and the corresponding functions $\phi_{w_1}$ and $\chi_{w_2}$. Accordingly, after discarding the constants, the {\em negative} loss \eqref{eq:empirical-symmetric-loss0} can be written as a function of the parameters
\begin{multline}
    {\cal L}(w_1, w_2; \I) = \frac{\sum\limits_{i} \Vert \chi_{w_2}( u_i - \phi_{w_1}( \chi_{w_2}, \uox, \I) ) \Vert^2}{\sum\limits_{i} \Vert \uix \Vert^2} \\
    + \frac{\sum\limits_{i} \Vert  (1-{\chi_{w_2}})( u_i - \phi_{w_1}( 1-{\chi_{w_2}}, \uix, \I) \Vert^2}{\sum\limits_{i} \Vert \uox \Vert^2}
    \label{eq:empirical-symmetric-loss}
\end{multline}
$\phi_{w_1}$ is called the {\em inpainter network}, and must be chosen to {\em minimize} the loss above. At the same time, the region $\o$, represented by the parameters $w_2$ of its indicator function $\chi_{w_2}$ called {\em mask generator network}, should be chosen so that $\uo$ is as uninformative as possible of $\ui$, and therefore the same loss is {\em maximized} with respect to $w_2$. This naturally gives rise to a minimax problem:
\begin{equation}
    \hat w = \arg\min_{w_1}\max_{w_2} {\cal L}(w_1, w_2; \I).
    \label{eq:minimax-training-loss}
\end{equation}
This loss has interesting connections to classical region-based segmentation, but with a twist as we discuss next.

\section{Related Work}
\label{sect-relations}

\def\x{x}

To understand the relation of our approach to classical methods, consider the simplest model for region-based segmentation \cite{chan2001active}
\begin{equation}
    L(\o, c_i, c_o) = \int_\o | \ui(\x) - c_i |^2 d\x + \int_{\o^c} | \uo(\x) - c_o |^2 d\x 
\end{equation}
typically combined with a regularizing term, for instance the length of the boundary of $\o$. This is a convex infinite-dimensional optimization problem that can be solved by numerically integrating a partial differential equation (PDE). The result enjoys significant robustness to noise, provided the underlying scene has piecewise constant radiance and is measured by image irradiance, to which it is related by a simple ``signal-plus-noise'' model. Not many scenes of interest have piecewise constant radiance, although this method has enjoyed a long career in medical image analysis. If we enrich the model by replacing the constants $c_i$ with smooth functions, $\phi_i(\x)$, we obtain the celebrated Mumford-Shah functional \cite{mumford1989optimal}, also optimized by integrating a PDE. Since smooth functions are an infinite-dimensional space, regularization is needed, which opens the Pandora box of regularization criteria, not to mention hyperparameters: Too much regularization and details are missed; too little and the model gets stuck in noise-induced minima. A modern version of this program would replace $\phi(\x)$ with a parametrized model $\phi_w(\x)$, for instance a deep neural network with weights $w$ pre-trained on a dataset ${\cal D}$. In this case, the loss is a function of $w$, with natural model complexity bounds. Evaluating $\phi_w$ at a point inside, $\x \in \o$, requires knowledge of the entire function $\u$ {\em inside} $\o$, which we indicate with $\phi_w(\x, \ui)$:
\begin{equation}
    \int_\o | \ui(\x) - \phi_{w}(x, \ui) |^2 d\x + \int_{\o^c} | 
    \uo(\x) - \phi_{w}(x, \uo) |^2 d\x.
\end{equation}
Here, a network can just map $\phi_w(x,\ui) = \ui$ providing a trivial solution, avoided by introducing (architectural or information) bottlenecks, akin to explicit regularizers. We turn the table around and use the outside to predict the inside and vice-versa:
\begin{equation}
       \int_\o | \ui(\x) - \phi_{w}(x, \uo) |^2 d\x + \int_{\o^c} | \uo(\x) - \phi_{w}(x, \ui) |^2 d\x 
\end{equation}
After normalization and discretization, this leads to our loss function \eqref{eq:empirical-symmetric-loss0}. The two regions compete: for one to grow, the other has to shrink. In this sense, our approach relates to region competition methods, and specifically Motion Competition \cite{cremers2005motion}, but also to adversarial training, since we can think of $\phi$ as the ``discriminator'' presented in a classification problem (GAN \cite{arjovsky2017wasserstein}), reflected in the loss function we use. This also relates to what is called ``self-supervised learning,'' a misnomer since there is no supervision, just a loss function that does not involve externally annotated data. Several variants of our approach can be constructed by using different norms, or correspondingly different models for the joint and marginal distributions \eqref{eq:probability model}.

More broadly, the ability to detect independently moving objects is primal, so there is a long history of motion-based segmentation, or moving object detection. Early attempts to explicitly model occlusions include the layer model \cite{Wang_1994} with piecewise affine regions, with computational complexity improvements using graph-based methods \cite{Jianbo_Shi} and variational inference~\cite{Cremers_2004,Brox_2006,Sun_2013,yang2015self} to jointly optimize for motion estimation and segmentation; \cite{Ochs_2014} use of long-term temporal consistency and color constancy, making however the optimization more difficult and sensitive to parameter choices. Similar ideas were applied to motion detection in crowds~\cite{Brostow}, traffic monitoring~\cite{Beymer} and medical image analysis~\cite{elnakib2011medical}. Our work also related to the literature on visual attention \cite{itti2000saliency,bylinskii2015towards}.

More recent data-driven methods~\cite{Tokmakov_2017_LVO,Tokmakov_2017_LMP,Cheng_2017,Song_2018_PDB} learn discriminative spatio-temporal features and differ mainly for the type of inputs and architectures. Inputs can be either image pairs \cite{Song_2018_PDB,Cheng_2017} or image plus dense optical flow \cite{Tokmakov_2017_LVO,Tokmakov_2017_LMP}. Architectures can be either time-independent \cite{Tokmakov_2017_LMP}, or with recurrent memory \cite{Tokmakov_2017_LVO,Song_2018_PDB}. Overall, those methods outperform traditional ones on benchmark datasets \cite{Ochs_2014,Perazzi_2016}, but at the cost of requiring a large amount of labeled training data and with evidence of poor generalization to previously unseen data.

It must be noted that, unlike in Machine Learning at large, it is customary in video object segmentation to call \emph{``unsupervised''} methods that {\em do} rely on massive amounts of manually annotated data, so long as they do not require manual annotation at run-time. We adopt the broader use of the term where unsupervised means that there is no supervision of any kind both at training and test time.

Like classical variational methods, our approach does not need any annotated training data. However, like modern learning methods, our approach learns a contextual model, which would be impossible to engineer given the complexity of image formation and scene dynamics.

\section{Experiments}
\label{sect-experiments}

We compare our approach to a set of state-of-the-art baselines on the task of video object segmentation to evaluate the accuracy of detection. We first present experiments on a controlled toy-example,
where the assumptions of our model are perfectly satisfied.
The aim of this experiment is to get a sense of the capabilities of the presented approach in ideal conditions.
In the second set of experiments, we evaluate the effectiveness of the proposed
model on three public, widely used datasets: Densely Annotated VIdeo Segmentation (DAVIS)~\cite{Perazzi_2016}, Freiburg-Berkeley Motion Segmentation (FBMS59)~\cite{Ochs_2014}, and SegTrackV2~\cite{Tsai_2010_segtrack}. Provided the high degree of appearance and resolution differences between them, these datasets represent a challenging benchmark for any moving object segmentation method. 
While the DAVIS dataset has always a single object per scene, FBMS and SegTrackV2 scenes can contain multiple objects per frame.
We show that our method not only outperforms the unsupervised approaches, but even edges out other supervised algorithms that, in contrast to ours, have access to a large amount of labeled data with precise manual segmentation at training time.
For quantitative evaluation, we employ the most common metric for video object segmentation , \emph{i.e.} the mean Jaccard score, a.k.a. intersection-over-union score, $\mathcal{J}$. Given space constraints, we add additional evaluation metrics in the appendix~\cite{yang2019unsupervised}.

\subsection{Implementation and Networks Details}
\label{sect-implementation}
\textbf{Generator, \emph{G}:} Depicted on the left of Fig.~\ref{fig:method}, the generator architecture is a shrunk version of SegNet~\cite{segnet_2017}.
Its encoder part consists of 5 convolutional layers each followed by batch normalization, reducing the input image to $\frac{1}{4}$ of its original dimensions. The encoder is followed by
a set of 4 atrous convolutions with increasing radius (2,4,8,16). The decoder part consists of 5 convolutional layers, that, with upsampling, generate an output with the same size of the input image. As in SegNet~\cite{segnet_2017}, a final softmax layer generates the probabilities for each pixel to be foreground or background. The generator input consists of an RGB image $I_t$ and the optical flow $u_{t:t+\delta T}$ between $I_t$ and $I_{t+\delta T}$, to introduce more variations in the optical flows conditioned on image $I_t$. At training time, $\delta T$ is randomly sampled from the uniform distribution $\mathcal{U}=[-5,5]$, with $\delta T \neq 0$.
The optical flow $u_{t:t+\delta T}$ is generated with the pretrained PWC network~\cite{sun2018pwc}, given its state-of-the-art accuracy and efficiency.
The generator network has a total of 3.4M parameters.

\textbf{Inpainter, \emph{I}:} We adapt the architecture of CPN~\cite{yang2018conditional} to build our inpainter network. Its structure is depicted on the right of Fig.~\ref{fig:method}.
The input to this network consists of the input image $I_t$ and the flow masked according to the generator output, $\chi u$, the latter concatenated with $\chi$, to make the inpainter aware of the region to look for context. Differently from the CPN, these two branches are balanced, and have the same number of parameters.
The encoded features are then concatenated and passed to the CPN decoder, that outputs an optical flow $\hat{u} = \phi(\chi, (1-\chi)\u, \I_t)$ of the same size of the input image, whose inside is going to be used for the difference between $\ui$ and the recovered flow inside. Similarly, we can run the same procedure for the complement part. Our inpainter network has a total of 1.5M parameters.

At test time, only the generator \emph{G} is used. Given $I_t$ and $u_{t:t+\delta T}$, it outputs a probability for each pixel to be foreground or background, $P_t(\delta T)$.
To encourage temporal consistency, we compute the temporal average:
\begin{equation}
\overline{P_t}=\sum_{\delta T=-5, \neq 0}^{\delta T=5} P_t(\delta T)
\label{eq:postprocess}
\end{equation}
The final mask $\chi$ is generated with a CRF \cite{krahenbuhl2011efficient} post-processing step on the final $\overline{P_t}$. More details about the post-processing can be found in the appendix.

\subsection{Experiments in Ideal Conditions}

Our method relies on basic, fundamental assumptions: \emph{The optical flow of the foreground and of the background are independent}. To get a sense of the capabilities of our approach in ideal conditions, we artificially produce datasets where this assumption is fully satisfied.
The datasets are generated as a modification of DAVIS2016~\cite{Perazzi_2016}, FMBS~\cite{Ochs_2014}, and SegTrackV2~\cite{Tsai_2010_segtrack}. While images are kept unchanged, ground truth masks are used to artificially perturb the optical flow generated by PWC~\cite{sun2018pwc} such that foreground and background are statistically independent. More specifically, a different (constant) optical flow field is sampled from a uniform distribution independently at each frame, and associated to the foreground and the background, respectively. More details about the generation of those datasets and the visual results can be found in the Appendix.
As it is possible to observe in Table~\ref{tab:toy_example}, our method reaches very high performance in all considered datasets.
This confirms the validity of our algorithm and that our loss function~\eqref{eq:minimax-training-loss} is a valid and tractable approximation of the functional~\eqref{eq:loss-object}.

\begin{table}
\centering
\begin{tabular}{ c | c c c}
    \toprule
      &  DAVIS~\cite{Perazzi_2016} & FBMS59~\cite{Ochs_2014} & SegTrackV2~\cite{Tsai_2010_segtrack} \\
    \midrule
    $\mathcal{J} \uparrow$ &  92.5 & 88.5 & 92.1 \\
    \bottomrule
\end{tabular}
\caption{\textbf{Performance under ideal conditions:} When the assumptions made by our model are fully satisfied, our approach can successfully detect moving objects.. Indeed, our model reaches near maximum Jaccard score in all considered datasets.} 
\label{tab:toy_example}
\end{table}

\begin{table*}
\centering
\footnotesize
\begin{tabular}{ c  c c c c c c c c c}
    \toprule
      &  \super{PDB}~\cite{Song_2018_PDB} & \super{FSEG}~\cite{Jain_2017_fseg} &  \super{LVO}~\cite{Tokmakov_2017_LVO}   & \unsuper{ARP}~\cite{Koh_2017_arp} & \unsuper{FTS}~\cite{Papazoglou_2013_fts} & \unsuper{NLC}~\cite{Faktor_2014_nlc} &
      \unsuper{SAGE}~\cite{Wenguan_Wang_2015_sage} & \unsuper{CUT}~\cite{Keuper_2015_cut} & \unsuper{Ours}\\
     \midrule
     DAVIS2016~\cite{Perazzi_2016} $\mathcal{J} \uparrow$ & \textbf{77.2}	& 70.7 & 	75.9 & \unsuper{\textbf{76.2}} & 55.8 & 55.1 & 42.6 & 55.2 &	 \unsuper{71.5} \\
     FBMS59~\cite{Ochs_2014} $\mathcal{J} \uparrow$ & \textbf{74.0} & 	68.4 	& 65.1 & 59.8 & 47.7 & 51.5 & \unsuper{61.2} &  57.2  &	 \unsuper{\textbf{63.6}} \\
     SegTrackV2~\cite{Tsai_2010_segtrack} $\mathcal{J}\uparrow$  & 60.9 & 61.4 &  57.3 & 57.2 & 47.8 &  \textbf{67.2} & 57.6 & 54.3 & \unsuper{62.0} \\
     DNN-Based & Yes & Yes & Yes  & No & No & No & No & No & Yes \\
     Pre-Training Required & Yes & Yes & Yes & No & No & No & No & No & No \\
  
    \bottomrule
\end{tabular}
\caption{\textbf{Moving Object Segmentation Benchmarks:} We compare our approach with $8$ different baselines on the task of moving object segmentation. 
In order to do so, we use three popular datasets, \emph{i.e.} DAVIS2016~\cite{Perazzi_2016},
FBMS59~\cite{Ochs_2014}, and SegTrackV2~\cite{Tsai_2010_segtrack}.
Methods in \super{blue} require ground truth annotations at training time and are pre-trained on image segmentation datasets.
In contrast, methods in \unsuper{red} are unsupervised and not require any ground-truth annotation.
Our approach is top-two in all the considered benchmarks, comparing to the other unsupervised methods. \textbf{Bold} indicates best among all methods, while \unsuper{\textbf{Bold Red}} and \unsuper{red} represent the best and second best for unsupervised methods, respectively.} 
\label{tab:all}
\end{table*}

\subsection{Performance on Video Object Segmentation}

As previously stated, we use the term \emph{Unsupervised} with a different meaning with respect to its definition in literature of video object segmentation. In our definition and for what follows, the supervision refers to the algorithm's usage of ground truth object annotations at training time.
In contrast, the literature usually defines methods as semi-supervised,
if at test time they assume the ground-truth segmentation of the first frame to be known~\cite{bao2018cnim, Man18pami}.
This could be posed as tracking problem~\cite{yang2015shape}
since the detection of the target is human generated.
Instead, here we focus on moving object detection and thus we compare our approach to the methods that are usually referred to as ``unsupervised'' in the video object segmentation domain. However we make further differentiation on whether the ground truth object segmentation is needed (supervised) or not (truly unsupervised) during training.

In this section we compare our method with other $8$ methods that represent the state of the art for moving object segmentation.
For comparison, we use the same metric defined above, which is the Jaccard score $\mathcal{J}$ between the real and predicted masks.

Table~\ref{tab:all} shows the performance of our method and the baseline methods on three popular datasets, DAVIS2016~\cite{Perazzi_2016}, FBMS59~\cite{Ochs_2014} and SegTrackV2~\cite{Tsai_2010_segtrack}.
Our approach is top-two in each of the considered datasets, and even outperforms baselines that need a large amount of labelled data at training time, \emph{i.e.} FSEG~\cite{Jain_2017_fseg}.

As can be observed in Table~\ref{tab:all}, unsupervised baselines typically perform well in one dataset but significantly worse in others.
For example, despite being the best performing unsupervised method on DAVIS2016, the performance of ARP~\cite{Koh_2017_arp} drops significantly in the FBMS59~\cite{Ochs_2014} and SegTrackV2~\cite{Ochs_2014} datasets.
ARP outperforms our method by 6.5\% on DAVIS, however, {\em our method outperforms ARP by 6.3\% and 8.4\%, on FBMS59 and SegTrackV2 respectively}.
Similarly, NLC~\cite{Faktor_2014_nlc} and SAGE~\cite{Wenguan_Wang_2015_sage} are extremely competitive in the Segtrack and FBMS59 benchmarks, respectively, but not in others.
NLC outperforms us on SegTrackV2 by 8.4\%, however {\em we outperform NLC by 29.8\% and 24.7\%, on DAVIS and FBMS respectively}.

It has been established that being second-best in multiple benchmarks is more indicative of robust performance than being best in one \cite{pang2013finding}.
Indeed, existing unsupervised approaches for moving object segmentation are typically highly-engineered pipeline methods which are tuned on one dataset but do not necessarily generalize to others. Also, consisting of several computationally intensive steps, extant unsupervised methods are generally orders of magnitude slower than our method (Table~\ref{tab:runtime_res}).

Interestingly, a similar pattern is observable for supervised methods.
This is particularly evident on the SegTrackV2 dataset~\cite{Tsai_2010_segtrack}, which is particularly challenging since several frames have very low resolution and are motion blurred.
%
Indeed, supervised methods have difficulties with the covariate shift due to changes in the distribution between training and testing data.
Generally, supervised methods alleviate this problem by pre-training on image segmentation datasets, but this solution clearly does not scale to every possible case.
In contrast, our method can be finetuned on any data without the need for the latter to be annotated.
As a result, our approach outperforms the majority of unsupervised methods as well as all the supervised ones, in terms of segmentation quality and training efficiency.

\subsection{Qualitative experiments and Failure Cases}

\begin{figure*}[!ht]
\def\mspace{0.1\linewidth}
  \centering
     \setlength{\tabcolsep}{13pt}
     \begin{tabular}{cccccccc}
         GT & SFL\cite{Cheng_2017_sfl} & LMP\cite{Tokmakov_2017_LMP} & PDB\cite{Song_2018_PDB} & CVOS\cite{Taylor_2015_cvos} & FTS\cite{Papazoglou_2013_fts} & ELM\cite{lao2018extending_elm} & Ours \\
    \end{tabular}
  \includegraphics[width=\textwidth]{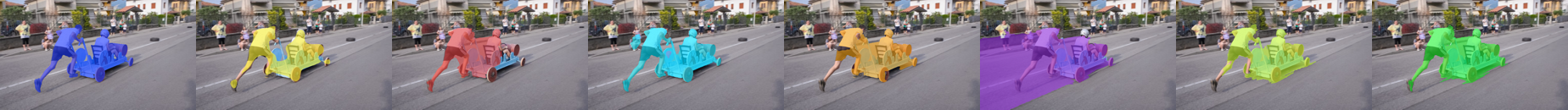}
  \includegraphics[width=\textwidth]{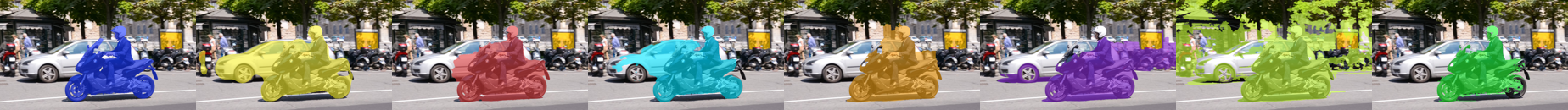}
  \includegraphics[width=\textwidth]{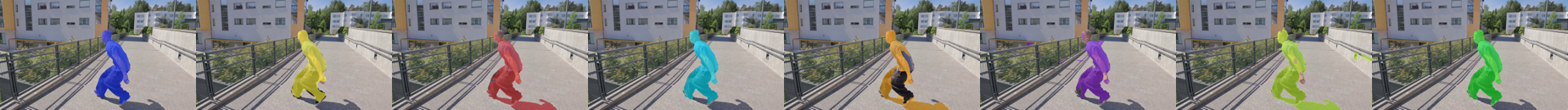}
  \includegraphics[width=\textwidth]{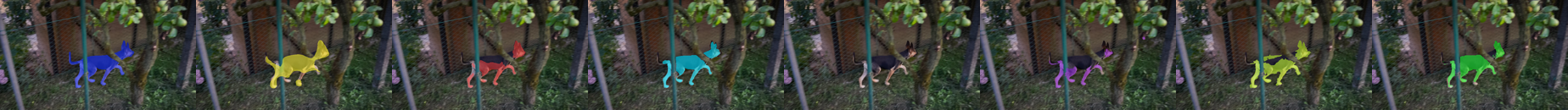}
  \includegraphics[width=\textwidth]{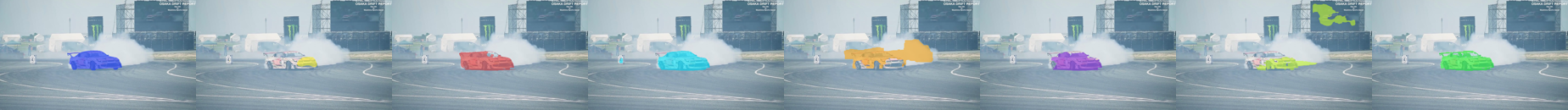}
  \includegraphics[width=\textwidth]{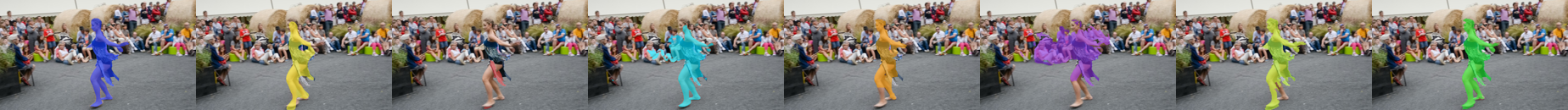}
  \includegraphics[width=\textwidth]{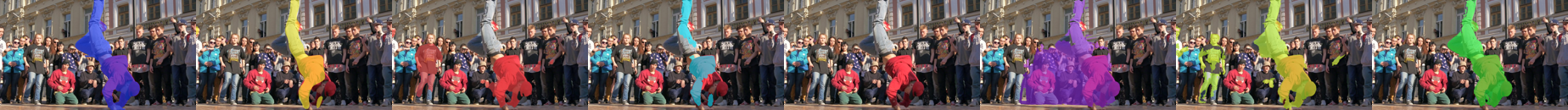}
  \includegraphics[width=\textwidth]{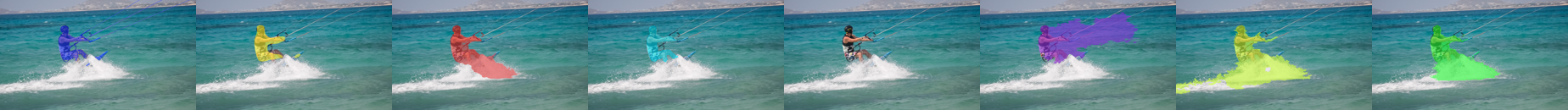}
  \caption{\textbf{Qualitative Results:} We qualitatively compare the performance of our approach with several state-of-the-art baselines as well as the Ground-Truth (GT) mask. Our prediction are robust to background clutter, large depth discontinuities and occlusions.  The last row shows a typical failure case of our method, \emph{i.e.} objects which are moved by the primary objects are detected as foreground (water is moved by the surfer in this case).}
  \label{fig:visual-comparisioin-davis2016}
\end{figure*}

In Fig.~\ref{fig:visual-comparisioin-davis2016} we show a qualitative comparison of the detection generated by our and others' methods on the DAVIS dataset.
Our algorithm can segment precisely the moving object regardless of cluttered background, occlusions, or large depth discontinuities.
The typical failure case of our method is the detection of objects whose motion is due to the primary object.
An example is given in the last row of Fig.~\ref{fig:visual-comparisioin-davis2016}, where the water moved by the surfer is also classified as foreground by our algorithm.

\begin{table*}[ht]
\centering
\begin{tabular}{ c  c c c c c c}
    \toprule
      & { \unsuper{ARP}~\cite{Koh_2017_arp}} & {\unsuper{FTS}~\cite{Papazoglou_2013_fts}} & { \unsuper{NLC}~\cite{Faktor_2014_nlc}} & {\unsuper{SAGE}~\cite{Wenguan_Wang_2015_sage}} & { \unsuper{CUT}~\cite{Keuper_2015_cut}} & { \unsuper{Ours}} \\
      \midrule
    {Runtime(s)} & {74.5} & {0.5} & {11.0} & { 0.88} & {103.0} & { \textbf{0.098}}\\
     { DNN-based} & { No} & {No} & {No} & {No} & {No} & {Yes} \\
    \bottomrule
\end{tabular}
\caption{\textbf{Run-time analysis}: Our method is not only effective (top-two in each considered dataset), but also orders of magnitude faster than other unsupervised methods. All timings are indicated without optical flow computation.} 
\label{tab:runtime_res}
\end{table*}

\subsection{Training and Runtime Analysis}

The generator and inpainter network's parameters are trained at the same time by minimizing the functional~\eqref{eq:minimax-training-loss}. The optimization time is approximately 6 hours on a single GPU Nvidia Titan XP 1080i. Since both our generator and inpainter networks are relatively small, we can afford very fast training/finetuning times. This stands in contrast to larger modules, \emph{e.g.} PDB~\cite{Song_2018_PDB}, that require up to 40 hrs of training.

At test time, predictions $\overline{P_t}$ (defined in eq.~\ref{eq:postprocess}) are generated at 3.15 FPS, or with an average time of 320ms per frame, including the time to compute optical flow with PWC~\cite{sun2018pwc}. Excluding the time to generate optical flow, our model can generate predictions at 10.2 FPS, or 98ms per frame.
All previous timings do not include the CRF post-processing step.
Table~\ref{tab:runtime_res} compares the inference time of our method with respect to other unsupervised methods.
Since our method at test time requires only a pass through a relatively shallow network, it is orders of magnitude faster than other unsupervised approaches.


\section{Discussion}
\label{sect-discussion}

Our definition of objects and the resulting inference criterion are related to generative model-based segmentation and region-based methods popular in the nineties.
However, there is an important difference: Instead of using the evidence inside a region to infer a model of that region which is as accurate as possible, we use evidence {\em everywhere else but} that region to infer a model within the region, and we seek the model to be as bad as possible.
This relation, explored in detail in Sect. \ref{sect-relations},  forces learning a contextual model of the image, which is not otherwise the outcome of a generative model in region-based segmentation. For instance, if we choose a rich enough model class, we can trivially model the appearance of an object inside an image region as the image itself. This is not an option in our model: We can only predict the inside of a region by looking outside of it. This frees us from having to impose modeling assumptions to avoid trivial solutions, but requires a much richer class of function to harvest contextual information. 

This naturally gives rise to an adversarial (min-max) optimization: An inpainter network, as a discriminator, tries to hallucinate the flow inside from the outside, with the reconstruction error as a quality measure of the generator network, which tries to force the inpainter network to do the lousiest possible job.


The strengths of our approach relate to its ability to learn complex relations between foreground and background without any annotation.
This is made possible by using modern deep neural network architectures like SegNet~\cite{segnet_2017} and CPN~\cite{yang2018conditional} as function approximators.
%


Not using ground-truth annotations can be seen as a strength but also a limitation: If massive datasets are available, why not use them? In part because even massive is not large enough: We have shown that models trained on large amount of data still suffer performance drops whenever tested on a new benchmark significantly different from the training ones. 
Moreover, our method does not require any pre-training on large image segmentation datasets, and it can adapt to any new data, since it does not require any supervision.
%
%
This adaptation ability is not only important for computer vision tasks, but can also benefit other applications, e.g. robotic navigation~\cite{dronet, deep_drive} or manipulation~\cite{khatib1985real}.

Another limitation of our approach is that, for the task of motion-based segmentation, we require the optical flow between subsequent frames. One could argue that optical flow is costly, local, and error-prone.
However, our method is general and could be applied to other statistics than optical flow. Such extensions are part of our future work agenda.
%
In addition, our approach does not fully exploit the intensity image, although we use it as a conditioning factor for the inpainter network.
An optical flow or an image can be ambiguous in some cases, but the combination of the two is rarely insufficient for recognition \cite{yang2015self}.
Again, our framework allows in theory exploitation of both, and in future work we intend to expand in this direction.


\section{Appendix}

{\bf Notation}

$\o \subset D$: $\o$ is a subset of the image domain $D$.\\
\indent $i \in \o$: pixel $i$ in region $\o$.\\
\indent $u_{\o}$: optical flow $u$ restricted to region $\o$.\\
\indent $\mathbb{I}(u_i, u_j | I)$: conditional mutual information between $u_i$ and $u_j$ given image $I$.

\subsection{More on the definition of objects}

Our definition Eq.~\eqref{eq:define-object} of object is in terms of mutual information between optical flows restricted to different regions. It is helpful to analyze some simple cases that follow directly from the definition.

\textbf{statement 1:} \emph{a subset of an object informs the remaining part of this object}. If the object is $\o$, and there is a subset $\hat{\o} \subset \o$, suppose $i \in \hat{\o}$, $j \in \o \setminus \hat{\o}$ respectively, then: $ \mathbb{I}(u_{\hat{\o}}, u_{\o \setminus \hat{\o}} | I) \geq \mathbb{I}(u_i, u_{\o \setminus \hat{\o}} | I) \geq \mathbb{I}(u_i, u_j | I) > 0 $ by Eq.~\eqref{eq:define-object}.

\textbf{statement 2:} \emph{a subset of the foreground does not inform a subset of the background}. Suppose $\o$ is the foreground, if $\hat{\o} \subset \o$, and $\o' \subset D \setminus \o$, then $\mathbb{I}(u_{\hat{\o}}, u_{\o'} |I) = 0$. Otherwise, we can find at least two pixels $i \in \hat{\o}$, $j \in \o'$ such that $\mathbb{I}(u_i, u_j |I) > 0$, which is contradictory to definition Eq.~\eqref{eq:define-object}.

\subsection{The optimality and uniqueness of objects}

In the main paper, we formalize the notion of foreground as a region $\o$ that is uninformed by the background, see Eq.~\eqref{eq:define-object}. Objects as we defined them in Eq.~\eqref{eq:define-object} are the regions that minimize the loss function Eq.~\eqref{eq:loss-object}.

\textbf{Proof:} First we show that the estimate $\o^*$ right on the object achieves the minimum value of the loss function, since:
\begin{multline}
    \mathcal{L}(\o^*; I) = \gamma(\o^* | D \setminus \o^*; I) + \gamma(D \setminus \o^*|\o^*; I) \\
    = \dfrac{\mathbb{I}(\u_{\o^*}, \u_{D \setminus \o^*} | I)}{\mathbb{H}(\u_{\o^*} | I)} + \dfrac{\mathbb{I}(\u_{D \setminus \o^*}, \u_{\o^*} | I)}{\mathbb{H}(\u_{D \setminus \o^*} | I)} = 0
\end{multline}
by statement (2) above. Thus $\o^*$ achieves the minimum value of the loss Eq.~\eqref{eq:loss-object}. Now we need to show that $\o^*$ is unique, for which, we just need to check the following two mutually exclusive and collectively inclusive cases for $\hat{\o} \neq \o^*$ (note that $\mathcal{L}(\emptyset; I)=\mathcal{L}(D; I)=1.0$ as $0 < \epsilon \ll 1$ is added to the denominator):
\begin{itemize}
    \item $\hat{\o}$ is either a subset of foreground or a subset of background: $\hat{\o} \cap D \setminus \o^* = \emptyset$ or $\hat{\o} \cap \o^* = \emptyset$.
    \item $\hat{\o}$ is neither a subset of foreground nor a subset of background: $\hat{\o} \cap D \setminus \o^* \neq \emptyset$ and $\hat{\o} \cap \o^* \neq \emptyset$.
\end{itemize}
In both cases $\mathcal{L}(\hat{\o}; I)$ is strictly larger than $0$ with some set operations under statements (1,2) above. Thus the object satisfies the definition Eq.~\eqref{eq:define-object} is a unique optima of the loss Eq.~\eqref{eq:loss-object}.

\subsection{Extra quantitative evaluations}

The performance of our algorithm compared with state-of-the-art systems on DAVIS2016~\cite{Perazzi_2016} can be found in Table~\ref{tab:davis_additional}.
The metrics used to perform the quantitative evaluation are the Jaccard score $\mathcal{J}$, and the mean boundary measure $\mathcal{F}$. For more details see~\cite{Perazzi_2016}. According to the same metrics, we also provide the per-category score for DAVIS and FBMS59 in Table~\ref{tab:davis_cat_additional} and ~\ref{tab:fbms_additional}.

\begin{table*}
\centering
\begin{tabular}{ c | c c c c c c c c c}
    \toprule
      &  \super{PDB}~\cite{Song_2018_PDB} &  \super{LVO}~\cite{Tokmakov_2017_LVO} & \super{FSEG}~\cite{Jain_2017_fseg} & \super{LMP}~\cite{Tokmakov_2017_LMP} & \unsuper{ARP}~\cite{Koh_2017_arp} & \unsuper{SFL}~\cite{Cheng_2017_sfl} & \unsuper{FTS}~\cite{Papazoglou_2013_fts} &
      \unsuper{NLC}~\cite{Faktor_2014_nlc} & \unsuper{Ours}\\
     \midrule
     $\mathcal{J}$ mean $\uparrow$ & 77.2 & 	75.9 & 70.7 & 	70.0 & 76.2 & 67.4 & 55.8 & 55.1 & 71.5 \\
$\mathcal{J}$ recall $\uparrow$ & 90.1	& 	89.1 	& 83.5 & 	85.0 &
91.1 & 81.4 &	64.9 & 55.8 & 86.5 \\
    $\mathcal{J}$ decay $\downarrow$ & 0.9	& 	0.0	& 1.5 & 	1.3 &
    7.0 & 6.2 &	0.0 & 12.6 & 9.5 \\
    \bottomrule
    $\mathcal{F}$ mean $\uparrow$ & 74.5	& 72.1	& 65.3 & 	65.9 & 70.6 & 66.7 &	51.1 & 52.3 & 70.5 \\
    $\mathcal{F}$ recall $\uparrow$ & 84.4	& 	83.4	& 73.8 & 79.2 & 83.5 & 77.1 &	51.6 & 51.9 & 83.5 \\
    $\mathcal{F}$ decay $\downarrow$ & -0.2	& 	1.3	& 1.8 & 1.5 & 7.9 & 5.1 &	2.9 & 11.4 & 7.0 \\
    \bottomrule
\end{tabular}
\caption{\textbf{Performance on DAVIS2016~\cite{Perazzi_2016}:} Methods in \super{blue} require ground truth annotations at training time, while the ones in \unsuper{red} are fully unsupervised. Our approach reaches comparable performance to the state-of-the-art, outperforming several supervised methods.}
\label{tab:davis_additional}
\end{table*}

\begin{table*}
\centering
\begin{tabular}{ c | c c c c c c}
    \toprule
    Category & $\mathcal{J}$ mean $\uparrow$ & $\mathcal{J}$ recall $\uparrow$ & $\mathcal{J}$ decay $\downarrow$ & 
    $\mathcal{F}$ mean $\uparrow$ & $\mathcal{F}$ recall $\uparrow$ & $\mathcal{F}$ decay $\downarrow$ \\
    \midrule
blackswan & 69.1 & 100.0 & 25.2 & 72.4 & 100.0 & 23.6 \\                                                                                          
bmx-trees & 59.2 & 74.4 & 12.6 & 84.1 & 100.0 & -7.4 \\                                                                                           
breakdance & 82.4 & 100.0 & -0.6 & 84.1 & 100.0 & -3.3 \\                                                                                         
camel & 83.0 & 100.0 & 4.6 & 83.0 & 100.0 & 0.7 \\                                                                                                
car-roundabout & 87.6 & 100.0 & -0.8 & 76.5 & 98.6 & -5.5 \\                                                                                      
car-shadow & 78.6 & 100.0 & 15.4 & 74.3 & 100.0 & 6.2 \\                                                                                          
cows & 85.4 & 100.0 & 0.6 & 79.6 & 98.0 & -1.3 \\                                                                                                 
dance-twirl & 79.0 & 95.5 & 1.2 & 82.4 & 100.0 & 9.9 \\                                                                                           
dog & 80.0 & 100.0 & 10.2 & 76.1 & 94.8 & 17.8 \\                                                                                                 
drift-chicane & 62.0 & 80.0 & 10.9 & 76.1 & 90.0 & 18.4 \\                                                                                        
drift-straight & 67.9 & 87.5 & 16.9 & 57.7 & 54.2 & 43.0 \\                                                                                       
goat & 26.9 & 17.0 & -17.7 & 34.4 & 17.0 & -14.7 \\                                                                                               
horsejump-high & 79.6 & 100.0 & 15.6 & 87.6 & 100.0 & 9.4 \\                                                                                      
kite-surf & 23.9 & 0.0 & 6.7 & 44.0 & 20.8 & -5.2 \\                                                                                              
libby & 76.5 & 91.5 & 19.1 & 92.2 & 100.0 & 2.7 \\                                                                                                
motocross-jump & 70.6 & 78.9 & -1.6 & 53.4 & 57.9 & -7.0 \\                                                                                       
paragliding-launch & 72.7 & 100.0 & 25.0 & 44.3 & 43.6 & 33.3 \\                                                                                  
parkour & 83.4 & 98.0 & 5.4 & 88.9 & 100.0 & 7.9 \\                                                                                               
scooter-black & 76.5 & 95.1 & -3.5 & 70.2 & 95.1 & -2.0 \\                                                                                        
soapbox & 77.0 & 95.9 & 16.7 & 74.5 & 100.0 & 18.0 \\
\midrule
Mean & 71.5 & 86.5 & 9.5 & 70.5 & 83.5 & 7.0 \\ 
\bottomrule
\end{tabular}
\caption{\textbf{Performance on DAVIS2016~\cite{Perazzi_2016}:} Per category performance on the DAVIS2016 dataset.} 
\label{tab:davis_cat_additional}
\end{table*}

\begin{table*}
\centering
\begin{tabular}{ c | c c c c c c}
    \toprule
    Category & $\mathcal{J}$ mean $\uparrow$ & $\mathcal{J}$ recall $\uparrow$ & $\mathcal{J}$ decay $\downarrow$ & 
    $\mathcal{F}$ mean $\uparrow$ & $\mathcal{F}$ recall $\uparrow$ & $\mathcal{F}$ decay $\downarrow$ \\
    \midrule
camel01 & 78.3 & 100.0 & 3.3 & 83.1 & 100.0 & 4.9 \\                                                                                              
cars1 & 61.4 & 100.0 & 0.0 & 37.5 & 0.0 & 0.0 \\                                                                                                  
cars10 & 31.1 & 0.0 & -9.4 & 22.9 & 0.0 & 0.8 \\                                                                                                  
cars4 & 83.6 & 100.0 & 7.6 & 79.5 & 100.0 & 11.6 \\                                                                                               
cars5 & 79.0 & 100.0 & 9.2 & 78.5 & 100.0 & 14.1 \\                                                                                               
cats01 & 90.0 & 100.0 & 5.1 & 90.0 & 100.0 & 19.2 \\                                                                                              
cats03 & 75.8 & 100.0 & -12.6 & 73.8 & 100.0 & 0.8 \\                                                                                             
cats06 & 61.3 & 81.2 & 27.3 & 75.6 & 87.5 & 17.5 \\                                                                                               
dogs01 & 73.4 & 77.8 & 35.9 & 74.3 & 88.9 & 21.3 \\                                                                                               
dogs02 & 73.8 & 90.0 & -9.7 & 74.4 & 90.0 & 10.9 \\                                                                                               
farm01 & 82.4 & 91.7 & -24.6 & 71.0 & 83.3 & -29.0 \\                                                                                             
giraffes01 & 38.5 & 46.7 & -25.4 & 44.2 & 33.3 & -3.0 \\                                                                                          
goats01 & 47.3 & 61.5 & 55.8 & 61.0 & 61.5 & 27.8 \\                                                                                              
horses02 & 63.1 & 81.8 & 18.3 & 67.3 & 90.9 & 36.1 \\                                                                                             
horses04 & 59.3 & 82.1 & 1.5 & 60.0 & 87.2 & -5.9 \\                                                                                              
horses05 & 41.0 & 28.6 & -10.3 & 38.7 & 19.0 & 11.6 \\                                                                                            
lion01 & 53.5 & 57.1 & 12.7 & 63.3 & 100.0 & -3.4 \\                                                                                              
marple12 & 68.6 & 100.0 & -15.6 & 58.6 & 83.3 & -0.9 \\                                                                                           
marple2 & 67.7 & 66.7 & 33.5 & 55.6 & 55.6 & 47.3 \\                                                                                              
marple4 & 68.2 & 100.0 & -3.9 & 58.9 & 100.0 & 3.0 \\                                                                                             
marple6 & 53.7 & 57.9 & 17.6 & 41.8 & 21.1 & 22.0 \\                                                                                              
marple7 & 58.1 & 78.6 & -1.8 & 35.6 & 21.4 & -17.5 \\                                                                                             
marple9 & 66.0 & 100.0 & 2.7 & 41.5 & 16.7 & -3.9 \\                                                                                              
people03 & 62.4 & 87.5 & 23.3 & 56.2 & 100.0 & 7.8 \\                                                                                             
people1 & 88.2 & 100.0 & 1.6 & 96.7 & 100.0 & 2.6 \\                                                                                              
people2 & 82.1 & 100.0 & -7.2 & 81.3 & 100.0 & -0.6 \\                                                                                            
rabbits02 & 50.2 & 66.7 & -5.9 & 55.5 & 75.0 & -4.9 \\                                                                                            
rabbits03 & 45.4 & 40.0 & -7.7 & 59.7 & 100.0 & -3.5 \\                                                                                           
rabbits04 & 44.0 & 50.0 & 31.2 & 50.9 & 57.1 & 26.6 \\                                                                                            
tennis & 70.9 & 100.0 & -4.9 & 78.1 & 100.0 & -2.4 \\  
\midrule
Mean & 63.6 & 78.2 & 4.9 & 62.2 & 72.4 &  7.02 \\
\bottomrule
\end{tabular}
\caption{\textbf{Performance on FBMS59~\cite{Ochs_2014}:} Per category performance on FBMS59.}
\label{tab:fbms_additional}
\end{table*}

\subsection{Details on CRF}

We use the online implementation\footnote{\url{https://github.com/lucasb-eyer/pydensecrf}} of the CRF algorithm \cite{krahenbuhl2011efficient} for post-processing the mask $\overline{P_t}$. We only use the pairwise bilateral potential with the parameters: sxy=25, srgb=5, compat=5 as defined in the corresponding function.

\subsection{Experiments in ideal conditions}
\label{sec:ideal_dataset_appx}

\begin{figure*}
        \centering
        \begin{tabular}{c c c c}
             \includegraphics[width=0.24\textwidth]{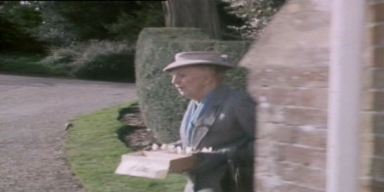} &  \includegraphics[width=0.24\textwidth]{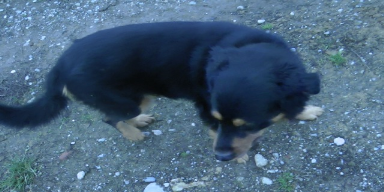} & \includegraphics[width=0.24\textwidth]{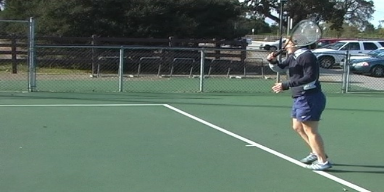} &
             \includegraphics[width=0.24\textwidth]{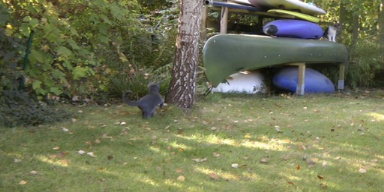}\\
            \includegraphics[width=0.24\textwidth]{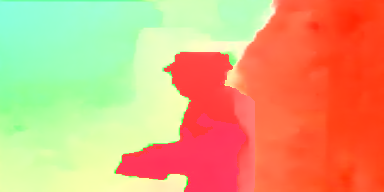} &   \includegraphics[width=0.24\textwidth]{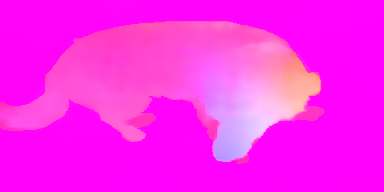} & \includegraphics[width=0.24\textwidth]{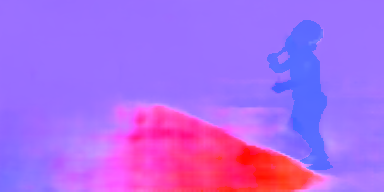} &
            \includegraphics[width=0.24\textwidth]{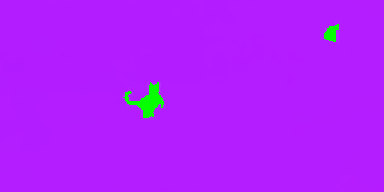}\\
              \includegraphics[width=0.24\textwidth]{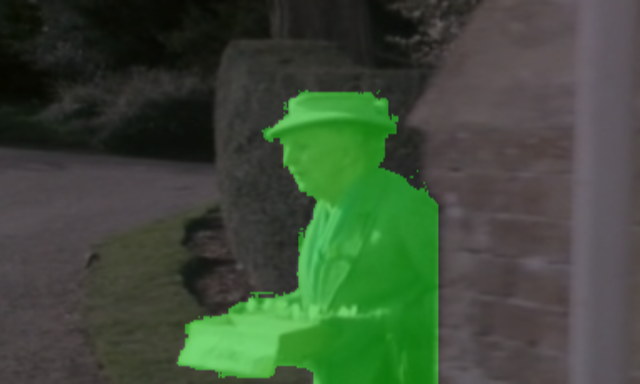} &  \includegraphics[width=0.24\textwidth]{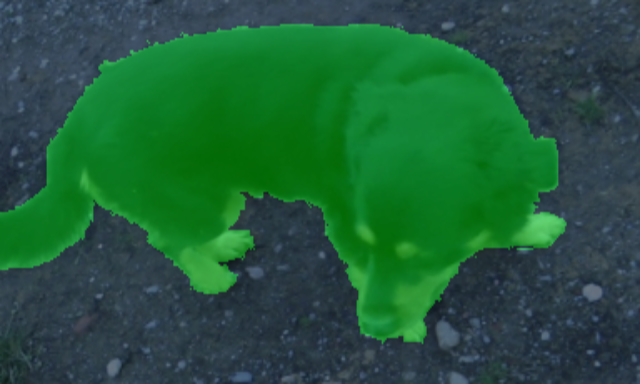} & \includegraphics[width=0.24\textwidth]{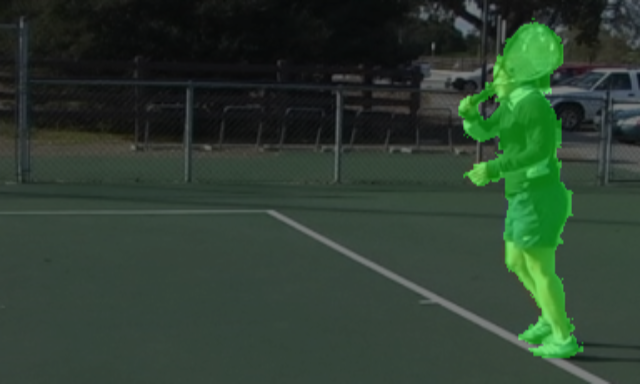} &
              \includegraphics[width=0.24\textwidth]{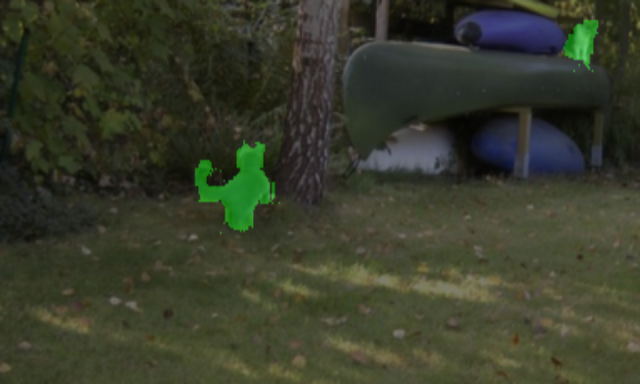}\\
        \end{tabular}
        \caption{\textbf{Experiment on FBMS in ideal conditions}: The first row shows some samples of input images, the second row their idealized optical flows, and the latter row the segmentation generated by our algorithm.}
        \label{tbl:table_of_figures_fbms}
\end{figure*}

\begin{figure*}
        \centering
        \begin{tabular}{c c c c}
             \includegraphics[width=0.24\textwidth]{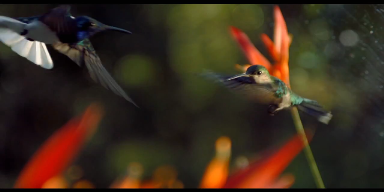} & \includegraphics[width=0.24\textwidth]{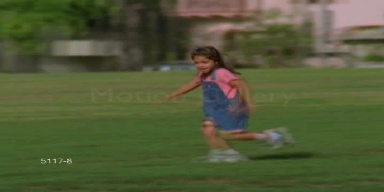} & \includegraphics[width=0.24\textwidth]{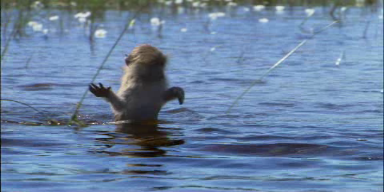} & \includegraphics[width=0.24\textwidth]{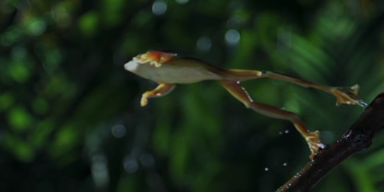} \\
            \includegraphics[width=0.24\textwidth]{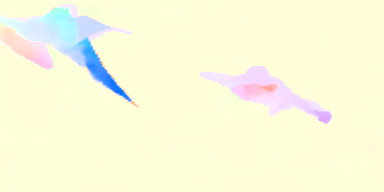} & \includegraphics[width=0.24\textwidth]{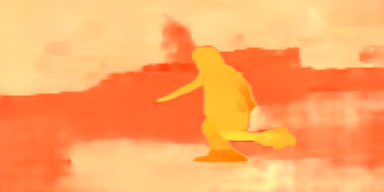} & \includegraphics[width=0.24\textwidth]{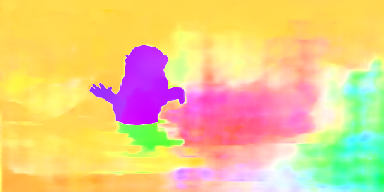} & \includegraphics[width=0.24\textwidth]{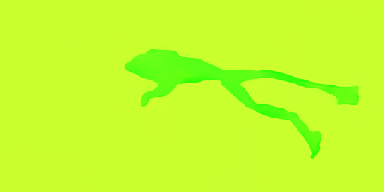} \\
              \includegraphics[width=0.24\textwidth]{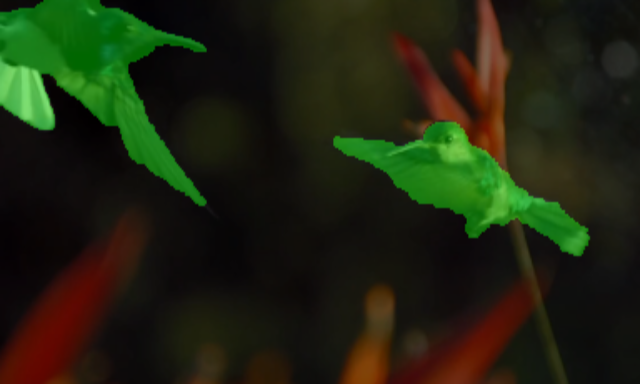} & \includegraphics[width=0.24\textwidth]{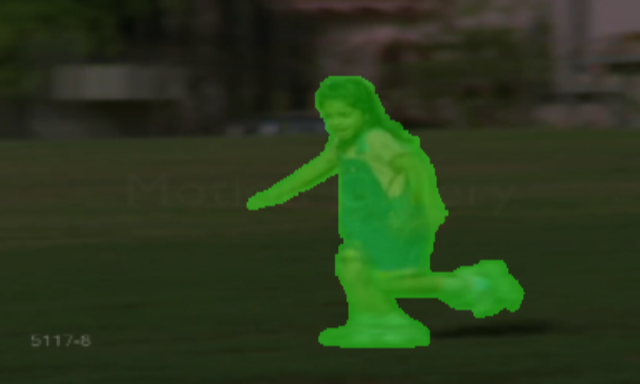} & \includegraphics[width=0.24\textwidth]{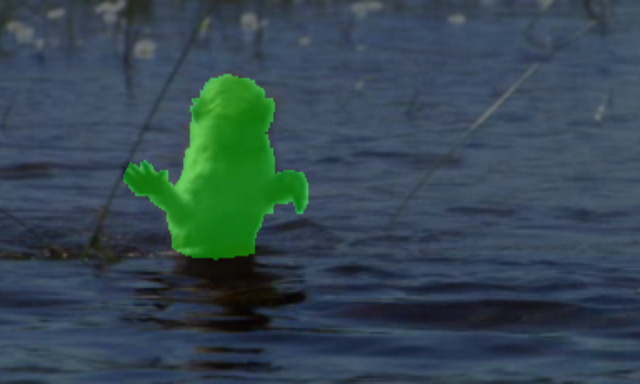} & \includegraphics[width=0.24\textwidth]{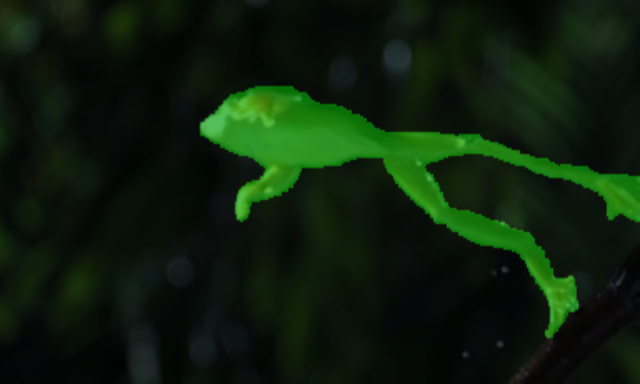} \\
        \end{tabular}
        \caption{\textbf{Experiment on SegTrackV2 in ideal conditions}: The first row shows samples of input images, the second row the idealized optical flows, and the latter row the segmentation generated by our algorithm.}
        \label{tbl:table_of_figures_segtrack}
\end{figure*}

The experiments under ideal conditions have been specifically designed to test the performance of our algorithm when its assumptions are fully satisfied. In particular, the derivation presented in Section 2 assumes that the foreground and the background motion are completely independent given the image. In real datasets, this conditions is not always true: during the generation of the benchmark datasets, many times the camera tracks the moving object, creating a relation between the two motions.
To analyze the performance of our method when background and foreground are \emph{completely independent}, we artificially modify our considered benchmark datasets.
To generate an ideal sample, we proceed as follows: We first take two image, $I_t$ and $I_{t+\delta T}$, with $\delta T\in\mathcal{U}[-5,5]$. Then, we generate the optical flow between the two frame $u_{t:t+\delta T}$ using PWC Net~\cite{sun2018pwc}. Now, using the ground-truth mask, we artificially add to the latter a random optical flow, different between the foreground and background.
The random optical flows are generated from a rotation $r\in\mathcal{U}[-1,1]$ radians and translations $t_x, t_y \in\mathcal{U}[-30,30]$ pixels. Since the ground-truth mask is used to generate the flow, in several case a solution is easy to find (last column of Fig.~\ref{tbl:table_of_figures_fbms} and Fig.~\ref{tbl:table_of_figures_segtrack}). However, as the original optical flow can be complex, it is not always possible to easily observe the mask in the flow (first three columns of Fig.~\ref{tbl:table_of_figures_fbms} and Fig.~\ref{tbl:table_of_figures_segtrack}). Nonetheless, since the background and foreground are fully independent, our algorithm can accurately segment the objects.

\newpage

{\small
\bibliographystyle{ieee}
\bibliography{egbib}
}

\end{document}